\documentclass[11pt]{article}
\usepackage[T1]{fontenc}
\usepackage[utf8]{inputenc}
\usepackage{lmodern}
\usepackage{microtype}

\usepackage[margin=1in]{geometry}
\usepackage{amsmath,amssymb}
\usepackage{graphicx}
\usepackage{booktabs}
\usepackage[numbers,sort&compress]{natbib}
\usepackage[hidelinks]{hyperref}
\hypersetup{pdftitle={Length Is Cheap, Competition Is Not: Attention Predicts Context Rot in Retrieval},pdfauthor={Yinuo Liu (ORCID 0009-0006-1114-4781), University of California San Diego},pdfkeywords={context rot, long-context language models, attention dilution, distractors, knowledge conflicts, needle in a haystack, retrieval-augmented generation},pdfsubject={Computation and Language}}
\usepackage{url}
\providecommand{\tightlist}{\setlength{\itemsep}{0pt}\setlength{\parskip}{0pt}}
\usepackage{orcidlink}
\usepackage{tikz}
\definecolor{badgegray}{HTML}{555555}
\definecolor{ghblack}{HTML}{181717}
\definecolor{hfyellow}{HTML}{FFD21E}
\definecolor{arxivred}{HTML}{B31B1B}
\newcommand{\badge}[5]{\href{#1}{\tikz[baseline=(L.base)]{%
  \node[fill=badgegray,text=white,inner xsep=4pt,text height=1.6ex,text depth=0.45ex,font=\sffamily\scriptsize\bfseries] (L) {#2};%
  \node[fill=#4,text=#5,inner xsep=4pt,text height=1.6ex,text depth=0.45ex,anchor=west,font=\sffamily\scriptsize] at (L.east) {#3};}}}

\title{Length Is Cheap, Competition Is Not: Attention Predicts Context Rot in Retrieval}
\author{%
  \textbf{Yinuo Liu}\,\orcidlink{0009-0006-1114-4781}\\[0.4em]
  \normalsize Department of Computer Science and Engineering\\
  \normalsize University of California San Diego, La Jolla, CA, USA\\[0.3em]
  \small ORCID: \href{https://orcid.org/0009-0006-1114-4781}{\texttt{0009-0006-1114-4781}}
}
\date{}

\begin{document}
\maketitle

\begin{abstract}
Why do language models fail on long inputs? The common answer is length. We show that, for retrieval, length is cheap and competition is not. On five open models (0.5B to 3B parameters, four families), neutral filler costs nothing: accuracy stays at 0.97 to 1.00, up to 32K tokens on Qwen2.5-1.5B. Statements in the format of the evidence, placed at the same positions, drop it to 0.11 to 0.21. A single statement that gives the queried key a rival value multiplies the loss by 9 to 110, and every additional rival adds to it. Inside the model, one quantity tracks the failure: the attention mass left on the evidence. Across 126,473 trials it explains up to 58\% of the variance in loss; log length explains at most 8\%. The masked-filler protocol behind the claim that length alone hurts reproduces on MMLU (11 to 18 points lost at 30K tokens), but fails a visible control in retrieval: masking costs 46 accuracy points at 32K tokens, while the same filler left visible costs none. In retrieval, context rot is a competition for attention. We release the probe and all 126K trial-level measurements.
\end{abstract}


\begin{center}
\badge{https://github.com/yliusci/length-is-cheap}{Code}{github.com/yliusci/length-is-cheap}{ghblack}{white}\quad
\badge{https://huggingface.co/datasets/yliusci/length-is-cheap}{Data}{huggingface.co/datasets/yliusci/length-is-cheap}{hfyellow}{black}
\end{center}

\hypertarget{introduction}{%
\section{Introduction}\label{introduction}}

Thirty-two thousand tokens of neutral text cost a 1.5B-parameter model
nothing. A few sentences that contradict the evidence cost it most of
its accuracy. This paper is about the gap between those two facts.

Language models now accept hundreds of thousands of tokens
\citep{chen2023extending, peng2023yarn, ding2024longrope, mohtashami2023landmark, wu2022memorizing, grattafiori2024llama, yang2024qwen25},
yet they use far fewer
\citep{hsieh2024ruler, an2024does, lu2024insights}. Retrieval accuracy
declines as inputs grow
\citep{liu2023lost, hsieh2024ruler, bai2023longbench, zhang2024inftybench, kuratov2024babilong, levy2024same},
a degradation now called \emph{context rot}
\citep{hong2025contextrot, xia2026diagnosing}. The prevailing
explanation is length itself: \citet{du2025context} hid the added tokens
from attention, kept their positions, and still measured a loss. If
length is the cause, the remedy is architectural, and every added token
is a tax.

Length, however, never grows alone. Every added token is also added
content, and in realistic inputs much of that content resembles what the
model is looking for. Softmax attention \citep{vaswani2017attention}
makes the distinction consequential. The weight on the evidence is

\[a^* = \frac{e^{s^*}}{e^{s^*} + \sum_{j \neq *} e^{s_j}},\]

where \(s^*\) is the attention score of the evidence and \(s_j\) the
scores of all other keys. A thousand dissimilar tokens with low \(s_j\)
add little to the denominator; a handful of near-copies of the evidence
can halve \(a^*\). If this is what happens inside the model, context rot
is a problem of competition, not of size, the remedy is to curate what
enters the context, and \(a^*\) should predict the loss better than
length does.

We test this prediction directly. In a key-value retrieval task, we vary
context length from 512 to 32K tokens, fill the same positions either
with neutral material or with statements that share the format of the
evidence, and measure on every trial both the log-probability of the
correct answer and the attention mass \(a^*\) that the model places on
the evidence. We study five open models from 0.5B to 3B parameters from
four families. Figure 1 summarizes the outcome: at equal length, the
loss follows what the context contains and how much attention the
evidence keeps, not how many tokens surround it.

Our contributions are:

\begin{enumerate}
\def\labelenumi{\arabic{enumi}.}
\tightlist
\item
  \textbf{Length is cheap; competition is not.} With neutral filler,
  accuracy stays between 0.97 and 1.00 on four of five models, up to 4K
  tokens on all four and up to 32K on Qwen2.5-1.5B. Filling the same
  positions with statements in the evidence template drives accuracy to
  0.11 to 0.21 at 8K to 32K tokens. The loss concentrates on statements
  that give the queried key a rival value: one such statement multiplies
  the loss by 9 to 110, and the loss grows with their number at fixed
  length.
\item
  \textbf{Attention mass on the evidence predicts the loss.} Across
  126,473 trials on five models, \(\log(1/a^*)\) explains 35 to 58\% of
  the variance in loss, against 2 to 8\% for log length, and the gap
  persists from 8K to 32K tokens.
\item
  \textbf{Masking is not a length baseline.} Replicating the
  masked-filler protocol of \citet{du2025context}, MMLU accuracy falls
  by 11 to 18 points at 30K masked tokens on Llama-3.2-3B and
  Qwen2.5-7B. In retrieval, the same masking costs 46 points of accuracy
  at 32K tokens while the identical filler left visible costs none.
\item
  \textbf{An open probe.} We release the harness, the attention probe,
  and all trial-level data: code at
  \url{https://github.com/yliusci/length-is-cheap}, data at
  \url{https://huggingface.co/datasets/yliusci/length-is-cheap}.
\end{enumerate}

\hypertarget{related-work}{%
\section{Related Work}\label{related-work}}

\textbf{Long-context degradation.} Accuracy falls as inputs grow, well
before the nominal context limit
\citep{an2023leval, shaham2023zeroscrolls, bai2024longbench, yen2024helmet, vodrahalli2024michelangelo, agarwal2024manyshot, zou2024manyshot, wang2026intelligence}.
\citet{liu2023lost} showed that retrieval depends on where evidence sits
\citep[see
also][]{hsieh2024found, zhang2024found, wang2024eliminating, peysakhovich2023attention, wu2025emergence, chowdhury2026lost};
RULER \citep{hsieh2024ruler} found effective lengths far below
advertised ones; NoLiMa \citep{modarressi2025nolima} showed that
removing lexical overlap between question and evidence accelerates the
decline. \citet{hong2025contextrot} named the phenomenon \emph{context
rot} across 18 frontier models. These studies vary length and content
together, so they cannot say which of the two is responsible.

\textbf{Knowledge conflicts.} When a context contains statements that
disagree, models often follow the wrong one
\citetext{\citealp{longpre2021entitybased}; \citealp{chen2022rich}; \citealp{xie2023adaptive}; \citealp{wu2024clasheval}; \citealp[see][
for a survey]{xu2024knowledge}}. Proactive interference, in which
earlier values of a key disrupt retrieval of the latest one, drives
accuracy log-linearly toward zero as updates accumulate
\citep{wang2025unable}. This literature treats conflicts as a
reliability problem, separate from length. We show that, in long-context
retrieval, rivals carry most of the loss that grows with length.

\textbf{Length or content?} Two lines of work give opposite answers.
\citet{du2025context} masked every distracting token from attention
while preserving position indices and still observed drops of at least
7.9\% at 30K tokens, concluding that length alone hurts. Conversely,
studies that hold length fixed and vary what fills it find that
semantically competing material dominates
\citep{bianchi2025hidden, dai2024deniahl}: hard distractors collapse
accuracy within the first 10\% of the context \citep{gao2026first},
competing passages shorten retention at constant passage count
\citetext{\citealp{repantis2026separating}; \citealp[see
also][]{cuconasu2024power}; \citealp{yoran2023making}}, and near-miss
negatives, not corpus size, bound retrieval at scale
\citep{lin2026beyond, li2025haystack}. Earlier, \citet{levy2024same} and
\citet{shi2023large} showed that similar padding or a single irrelevant
sentence suffices to degrade reasoning in short contexts
\citep{jia2017adversarial, yang2025llm, wang2025adaptive}. Only
\citet{gao2026first} inspect attention, through logit margins on
retrieval heads; none relates the attention on the evidence to the loss
trial by trial, so none can tell how much of the failure competitors
explain by drawing attention away from the evidence.

\textbf{Attention dilution theory.} Softmax attention assigns each key a
share that shrinks as keys accumulate. \citet{chiang2022overcoming} and
\citet{velickovic2024softmax} proved that this dispersion limits sharp
retrieval as length grows
\citep{hahn2019theoretical, mudarisov2025limitations, vasylenko2025longcontext};
\citet{nakanishi2025scalablesoftmax} and \citet{chen2025critical}
derived length-dependent temperature corrections
\citep{li2025information, anson2025scaleinvariant};
\citet{ghaffari2026context} modeled distractors as extreme-value
competitors whose maximum score grows with their number, requiring the
evidence margin to grow as \(\sqrt{\log N}\).
\citet{tworkowski2023focused} first described this \emph{distraction
issue}: relevant keys become a vanishing fraction of all keys, and
similar keys compete. These accounts predict that degradation should
track the attention mass left on the evidence. That prediction has not
been tested trial by trial on measured attention in pretrained models.

\textbf{Our position.} We do not claim the phenomenon, the distractor
effect, or the dilution theory. We close the loop between them,
measuring behavior, content and attention on the evidence in the same
trials (Table 1). \citet{du2025context} also used a visible baseline,
but filled it with whitespace, which we find harmful on MMLU as they
did; with neutral text instead, length is free in retrieval. Their
masked condition, which we replicate, is the only evidence for a pure
length effect that remains, and we show that it fails a visible control.

\begin{table}[t]
\centering
\small
\caption{Controls in the closest prior work.}
\resizebox{\ifdim\width>\linewidth\linewidth\else\width\fi}{!}{\begin{tabular}{lcccc}
\toprule
Study & Length fixed, content varied & Visible neutral filler & Masked filler & Attention vs. loss, per trial \\
\midrule
\citet{du2025context} & no & whitespace & yes & no \\
\citet{gao2026first} & yes & repeated text & no & head-level margins \\
\citet{repantis2026separating} & yes & no & no & no \\
\citet{hong2025contextrot} & partial & no & no & no \\
\textbf{Ours} & \textbf{yes} & \textbf{three text types} & \textbf{yes, replicated} & \textbf{yes, 126K trials} \\
\bottomrule
\end{tabular}}
\end{table}

\hypertarget{setup}{%
\section{Setup}\label{setup}}

One principle governs the design: every comparison holds length and
evidence position fixed and changes only what surrounds the evidence. We
call this the \emph{visible-control protocol}. It needs forward passes
only and applies to any open-weight model.

\textbf{Task.} Each context hides one \emph{evidence} statement, ``The
secret code of the river is 4821.'', where the key (here, river) is a
common noun and the value a four-digit number. The context ends with the
query ``Question: what is the secret code of the river? Answer: The
secret code of the river is'', and we score the log-probability \(\ell\)
of the correct value under teacher forcing, summed over its tokens.
Tokenizers split the value differently (five tokens for Qwen and
SmolLM2, three for OLMo and Llama), so we compare losses within models,
not across them. The evidence sits at a depth drawn uniformly over the
context.

\textbf{Conditions.} For each trial we build four contexts that share
the evidence, the query, and the total length \(L\):

\begin{center}
\small
\resizebox{\ifdim\width>\linewidth\linewidth\else\width\fi}{!}{\begin{tabular}{ll}
\toprule
Condition & Content around the evidence \\
\midrule
\emph{short} & nothing \\
\emph{neutral} & random numbers, random words, or one repeated sentence \\
\emph{competing} & statements in the evidence template, keys drawn uniformly from 51 nouns \\
\emph{masked} & filler hidden from attention, positions preserved \\
\bottomrule
\end{tabular}}
\end{center}

We define the loss of a condition \(c\) as
\(\Delta_c = \log\left(p_{\text{short}} / p_c\right)\), in nats, where
\(p_c = e^{\ell_c}\) is the probability of the correct value in
condition \(c\). The \emph{neutral} condition isolates length; the
difference between \emph{competing} and \emph{neutral} isolates content
at fixed length.

Because keys are drawn uniformly, some competing statements share the
queried key and assign it a different value. These \emph{conflicts} are
the strongest competitors the filler contains. Their number
\(n_{\text{conf}}\) grows with length (mean 0.8 at 512 tokens, 3.2 at
2,048, 50.5 at 32,768; none in 42 to 48\% of 512-token contexts and in
no context of 4,096 tokens or more). We recover \(n_{\text{conf}}\) for
every trial by regenerating its context from the recorded seed, which
lets us separate competitors that share the key from those that only
share the format. Separately, we insert 0 to 5 explicit distractors that
share the format but never the key.

\textbf{Masking.} In \emph{masked}, a 2D attention mask removes every
filler token from attention for all subsequent queries while position
indices keep counting through them, so the evidence stays at its
original distance from the query and every competitor becomes invisible.
The first tokens of the sequence remain visible, preserving the
attention sink
\citep{xiao2023efficient, gu2024attention, sun2024massive, barbero2025llms, ranmilo2026attention, rizwan2026new, su2026attention}.

\textbf{Models and lengths.} We test Qwen2.5-0.5B-Instruct and
Qwen2.5-1.5B-Instruct \citep{yang2024qwen25}, SmolLM2-1.7B-Instruct
\citep{allal2025smollm2}, OLMo-2-0425-1B-Instruct \citep{walsh2024olmo}
and the base Llama-3.2-3B \citep{grattafiori2024llama}, at
\(L \in \{512, 1024, 2048\}\) tokens for all models and up to 32,768
tokens within each model's context window. All runs use float16 weights
and PyTorch SDPA attention, on one Apple workstation (64 GB unified
memory) and one RTX 4090 laptop GPU (16 GB).

\textbf{Attention mass on the evidence.} For each configuration we
select, in the \emph{short} context, the eight (layer, head) pairs that
place the most attention on the evidence tokens
\citep{olsson2022incontext, wu2024retrieval, zhang2025queryfocused}. In
every condition, \(a^*\) is the attention that these heads send from the
final query token to the evidence tokens, summed over evidence tokens
and averaged over heads. We do not renormalize \(a^*\) to exclude the
attention sink. We compare predictors of \(\Delta\) by the coefficient
of determination \(R^2\) of univariate least-squares fits on \(a^*\),
\(\log(1/a^*)\) and \(\log L\).

\textbf{Replication of \citet{du2025context}.} We reproduce their
masked-filler protocol on 1,000 MMLU questions with the base
Llama-3.2-3B and Qwen2.5-7B-Instruct, inserting 3,750, 7,500, 15,000 or
30,000 masked tokens between the question and its answer options,
following their layout, in which the masked positions separate the
evidence from the question. We implement masking by deleting the filler
tokens and advancing the position indices by the same amount, which we
verified to match explicit attention masking on Llama. We report
accuracy with Wilson 95\% intervals and paired McNemar tests against the
unpadded baseline.

\hypertarget{results}{%
\section{Results}\label{results}}

One comparison condenses the paper. On Qwen2.5-1.5B at 32K tokens, with
length and evidence position fixed, the median cost of neutral numeric
text is 0.004 nats, the cost of masking it is 1.20 nats, and the cost of
competing statements is 4.76 nats.

\hypertarget{length-is-cheap}{%
\subsection{Length is cheap}\label{length-is-cheap}}

Visible neutral filler barely moves the target. On Qwen2.5-1.5B, the
median loss stays at or below 0.006 nats at every length from 8K to 32K
tokens and for all three neutral fillers, and accuracy stays between
0.96 and 1.00 (Table 2). Qwen2.5-0.5B, SmolLM2 and Llama-3.2-3B show the
same behavior up to 4K tokens. OLMo-2-1B is the exception: it loses 0.57
nats with neutral filler at 2,048 tokens (accuracy 0.91).

\begin{figure}[t]
\centering
\includegraphics[width=\textwidth]{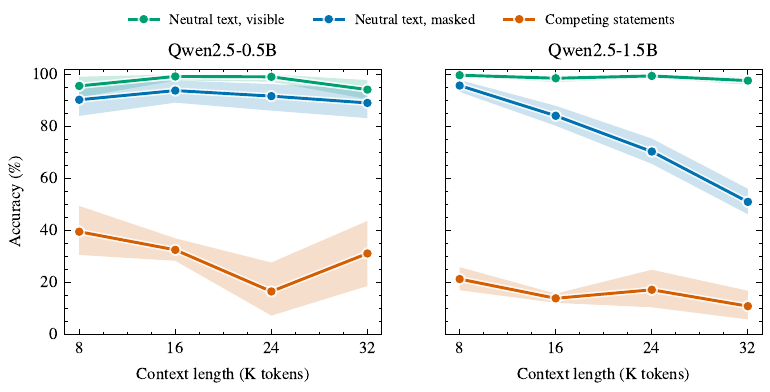}
\caption{Retrieval accuracy against context length, with length and evidence position fixed. Neutral text left visible (green) costs almost nothing; the same text masked at preserved positions (blue) costs up to 47 points on Qwen2.5-1.5B; competing statements (orange) cost 56 to 87 points at every length. Mean over trials, pooled over the three neutral fillers; shaded bands are 95\% bootstrap intervals.}
\end{figure}

\begin{table}[t]
\centering
\small
\caption{Qwen2.5-1.5B, neutral filler (numbers). Accuracy with the filler visible and with the same filler masked at preserved positions; median masked loss $\Delta_{\text{masked}}$ in nats.}
\resizebox{\ifdim\width>\linewidth\linewidth\else\width\fi}{!}{\begin{tabular}{lcccc}
\toprule
$L$ & 8K & 16K & 24K & 32K \\
\midrule
Accuracy, visible & 1.00 & 0.99 & 1.00 & 0.98 \\
Accuracy, masked & 0.93 & 0.87 & 0.73 & 0.52 \\
$\Delta_{\text{masked}}$ & 0.03 & 0.11 & 0.44 & 1.20 \\
\bottomrule
\end{tabular}}
\end{table}

\hypertarget{competition-is-not}{%
\subsection{Competition is not}\label{competition-is-not}}

Filling the same positions with competing statements instead of neutral
text turns a free context into a failing one. On Qwen2.5-1.5B, the
median loss reaches 2.82, 3.94, 4.03 and 4.76 nats at 8K, 16K, 24K and
32K tokens, and accuracy falls to 0.11 to 0.21; on Qwen2.5-0.5B, the
loss reaches 1.79 to 4.07 nats and accuracy 0.17 to 0.40. Length is the
same in both conditions; only content differs (Figure 1).

The loss concentrates on conflicts (Table 3). Across five models,
contexts without any statement that shares the queried key lose 0.02 to
0.34 nats; contexts with at least one lose 1.38 to 3.24 nats. Accuracy
is 0.97 to 1.00 without conflict and 0.25 to 0.52 with one. At fixed
length the loss grows with the number of conflicts: on Qwen2.5-1.5B at
2,048 tokens it is 0.03, 1.37, 2.24, 2.48, 2.71, 3.26 and 3.45 nats for
0 to 6 or more. Competitors that share only the format matter far less:
the number of explicit distractors explains at most 0.3\% of the
variance in loss (\(R^2 \le 0.003\)).

\begin{table}[t]
\centering
\small
\caption{Mean loss (nats) by presence of a same-key conflict, $L \le 2{,}048$.}
\resizebox{\ifdim\width>\linewidth\linewidth\else\width\fi}{!}{\begin{tabular}{lcc}
\toprule
Model & No conflict & At least one \\
\midrule
Qwen2.5-0.5B & 0.175 & 1.812 \\
Qwen2.5-1.5B & 0.020 & 2.238 \\
SmolLM2-1.7B & 0.022 & 1.381 \\
OLMo-2-1B & 0.058 & 2.796 \\
Llama-3.2-3B & 0.344 & 3.239 \\
\bottomrule
\end{tabular}}
\end{table}

\hypertarget{one-quantity-predicts-the-failure-attention-on-the-evidence}{%
\subsection{One quantity predicts the failure: attention on the
evidence}\label{one-quantity-predicts-the-failure-attention-on-the-evidence}}

Across 126,473 trials, \(\log(1/a^*)\) predicts the loss on every model,
and log length barely does (Table 4). \(R^2(\log(1/a^*))\) ranges from
0.35 to 0.58, against 0.02 to 0.08 for \(\log L\); a bootstrap
confidence interval on the difference excludes zero for each model, and
the ordering holds within each (length, depth) cell. Attention on the
evidence collapses from 0.51 to 0.77 in the short context to 0.17 to
0.31 with competing filler. The relation extends to long contexts: from
8K to 32K tokens, \(R^2(\log(1/a^*))\) is 0.32 on Qwen2.5-1.5B and 0.33
on Llama-3.2-3B, against 0.02 and 0.03 for \(\log L\); on SmolLM2-1.7B,
up to its 8K window, it is 0.45 against 0.18.

\begin{figure}[t]
\centering
\includegraphics[width=0.55\textwidth]{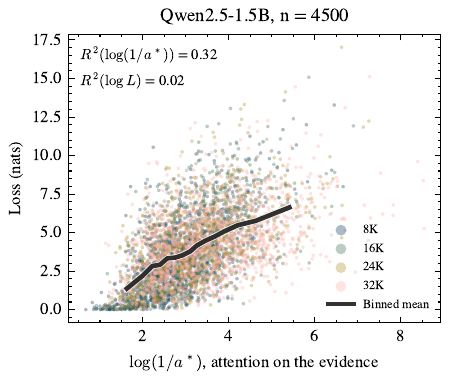}
\caption{Loss $\Delta$ (nats) against $\log(1/a^*)$ for 4,500 trials of Qwen2.5-1.5B with competing filler at 8K to 32K tokens, colored by length. Black: mean loss in 15 quantile bins of $\log(1/a^*)$. Lengths overlap completely, while attention orders the loss.}
\end{figure}

\begin{table}[t]
\centering
\small
\caption{Predictors of the loss, $L \in \{512, 1024, 2048\}$. Slope of $\Delta$ on $\log(1/a^*)$ with 95\% CI.}
\resizebox{\ifdim\width>\linewidth\linewidth\else\width\fi}{!}{\begin{tabular}{lccccc}
\toprule
Model & $n$ & $R^2(a^*)$ & $R^2(\log(1/a^*))$ & $R^2(\log L)$ & Slope \\
\midrule
Qwen2.5-0.5B & 32,318 & 0.23 & 0.35 & 0.03 & 1.40 [1.38, 1.42] \\
Qwen2.5-1.5B & 40,632 & 0.43 & 0.58 & 0.05 & 2.22 [2.21, 2.24] \\
SmolLM2-1.7B & 9,325 & 0.30 & 0.41 & 0.07 & 1.68 [1.64, 1.72] \\
OLMo-2-1B & 42,005 & 0.26 & 0.35 & 0.02 & 3.97 [3.92, 4.02] \\
Llama-3.2-3B & 2,193 & 0.36 & 0.45 & 0.08 & 2.75 [2.63, 2.88] \\
\bottomrule
\end{tabular}}
\end{table}

\(a^*\) summarizes the effect of conflicts rather than adding to it:
once \(\log(1/a^*)\) is known, \(n_{\text{conf}}\) raises \(R^2\) by at
most 0.04. Within contexts free of conflicts, where losses are small,
\(\log(1/a^*)\) explains only 2 to 10\% of the variance. The relation
has a threshold: on Qwen2.5-1.5B the loss stays near zero until \(a^*\)
falls to about 0.35, then rises steeply. The model tolerates losing
about half of the attention it gives the evidence in a short context,
then fails abruptly. Slopes exceed 1 on every model and differ across
models, so we report no universal law; the loss grows faster than the
removal of evidence mass alone would imply under a single-head softmax
reading.

\hypertarget{masking-is-not-a-length-control}{%
\subsection{Masking is not a length
control}\label{masking-is-not-a-length-control}}

Masking reproduces the length-alone effect of \citet{du2025context}. On
MMLU, accuracy falls with the number of masked tokens on both models,
plateauing beyond 15,000 on Llama-3.2-3B (Table 5; all paired McNemar
tests against the baseline \(p \le 0.007\)). From 7,500 tokens onward,
Qwen2.5-7B loses 11.3 to 17.7 points, against 15.5 to 21.1 for
Llama-3.1-8B in \citet[Table 3]{du2025context}, so the effect is not
specific to one model family.

\begin{table}[t]
\centering
\small
\caption{MMLU accuracy (\%) with masked filler, $n = 1{,}000$ per cell, Wilson 95\% intervals.}
\resizebox{\ifdim\width>\linewidth\linewidth\else\width\fi}{!}{\begin{tabular}{lccccc}
\toprule
Masked tokens & 0 & 3,750 & 7,500 & 15,000 & 30,000 \\
\midrule
Llama-3.2-3B & 51.7 [48.6, 54.8] & 44.7 [41.6, 47.8] & 42.4 [39.4, 45.5] & 39.3 [36.3, 42.4] & 40.5 [37.5, 43.6] \\
Qwen2.5-7B & 65.8 [62.8, 68.7] & 62.6 [59.6, 65.5] & 54.5 [51.4, 57.6] & 49.4 [46.3, 52.5] & 48.1 [45.0, 51.2] \\
\bottomrule
\end{tabular}}
\end{table}

The same intervention degrades retrieval, whereas the filler it hides
does not. In Table 2, the masked context loses 1.20 nats and 46 points
of accuracy at 32K tokens, while the identical filler left visible loses
nothing. Across the three neutral fillers at 32K, accuracy is 0.96 to
0.99 when visible and 0.48 to 0.54 when masked. A context of masked
tokens at preserved positions is therefore not a context of irrelevant
tokens: it presents the model with long positional gaps that it never
encounters in training
\citep{su2021roformer, press2021train, kazemnejad2023impact, wang2024length, an2024does, du2026rope},
and its cost should not be read as the cost of length. The effect is
model-dependent: on Qwen2.5-0.5B, masking costs 0.1 to 0.4 nats.

MMLU shows the same ordering at 4,096 tokens on Qwen2.5-1.5B: 59.1\%
without filler, 57.5\% with visible neutral filler, 53.9\% with the
filler masked. Length is not free in every form, however: visible
whitespace padding lowers the same model to 36.0\% at 4,096 tokens, an
effect absent at 2,048. We have no visible-filler arm on MMLU beyond 4K
tokens, so the drops in Table 5 cannot be attributed to the positional
gap alone.

\hypertarget{discussion}{%
\section{Discussion}\label{discussion}}

\textbf{Curate, do not truncate.} In retrieval, thousands of neutral
tokens cost nothing, while a few statements that compete with the
evidence cost most of the accuracy. The effective budget of a context is
therefore not its token count but the number of statements that compete
for the query. For retrieval-augmented generation
\citep{lewis2020retrievalaugmented, shim2025beyond}, agent memory
\citep{dongre2026attention} and long-document pipelines, this reverses a
common priority: deduplicating and resolving conflicting statements
before they reach the model
\citep{chirkova2025provence, wu2024reducing, choi2025think, bui2026make, arjmandi2026distractoraware}
should matter more than shortening the context.

\textbf{Measure attention, not only accuracy.} Conflicts, length and
depth reach the loss through one measurable quantity: once
\(\log(1/a^*)\) is known, the number of conflicts adds at most 0.04 to
\(R^2\). Softmax dilution accounts predict exactly this
\citep{tworkowski2023focused, ghaffari2026context, gu2026long}; we
observe it on measured attention, trial by trial. Because \(a^*\)
requires only a forward pass, it can serve as a diagnostic for any open
model and any retrieval benchmark. The threshold (near
\(a^* \approx 0.35\) on Qwen2.5-1.5B, at a model-dependent point
elsewhere) and slopes above 1 indicate a nonlinear readout: the model
absorbs a moderate loss of evidence mass, then fails abruptly. We
predict that methods which sharpen attention at long range
\citep{nakanishi2025scalablesoftmax, chen2025critical, lee2025seal, ye2026dysco, bansal2025not, pei2026focusft, zhang2024selective, leviathan2024selective, ye2024differential, qiu2025gated, huang2026threshold, lin2025forgetting, tan2024scaling}
help most when they keep \(a^*\) above it.

\textbf{Report a visible control.} Masking tokens at preserved positions
creates long positional gaps absent from pretraining
\citep{xiao2025mitigating, yang2025rope, gelberg2025extending}. In
retrieval, this intervention costs 46 accuracy points where the same
filler left visible costs none, so its cost is not the cost of length.
We recommend that any claim about the effect of length include a visible
neutral-text control at the same positions. Whitespace padding on MMLU
shows that some forms of length do hurt reasoning, so the two accounts
are not exclusive: competition dominates retrieval, and length effects
remain possible in reasoning.

\hypertarget{limitations}{%
\section{Limitations}\label{limitations}}

\(a^*\) is measured in the same forward pass as the loss and we ran no
intervention on it, so our evidence is predictive, not causal. The
predictor analysis covers 512 to 2,048 tokens on five models and 8K to
32K on two. In our competing filler, the number of conflicts grows with
length, and every context of 4,096 tokens or more contains at least one,
so the effect of format-only competitors above 2K tokens is not
measured; the conflict analysis was not preregistered. \(a^*\) is read
at a single query token, from eight heads, without removing the
attention sink. Value tokenization differs across models, so losses are
comparable within but not across models. All results come from one
synthetic retrieval task and models of at most 3B parameters, except the
7B model in the MMLU replication, which has no visible-filler arm beyond
4K tokens.

\hypertarget{conclusion}{%
\section{Conclusion}\label{conclusion}}

In retrieval, long contexts do not fail because they are long. They fail
when what they contain competes with the evidence, and the attention
that the evidence keeps tells us when. Context engineering, not context
length, is the lever.

\bibliographystyle{unsrtnat}
\bibliography{refs}

@article{vaswani2017attention,
  author={Vaswani, Ashish and Shazeer, Noam and Parmar, Niki and Uszkoreit, Jakob and Jones, Llion and Gomez, Aidan N. and others},
  title={{Attention Is All You Need}},
  journal={arXiv preprint arXiv:1706.03762},
  year={2017}
}

@article{su2021roformer,
  author={Su, Jianlin and Lu, Yu and Pan, Shengfeng and Murtadha, Ahmed and Wen, Bo and Liu, Yunfeng},
  title={{RoFormer: Enhanced Transformer with Rotary Position Embedding}},
  journal={arXiv preprint arXiv:2104.09864},
  year={2021}
}

@article{press2021train,
  author={Press, Ofir and Smith, Noah A. and Lewis, Mike},
  title={{Train Short, Test Long: Attention with Linear Biases Enables Input Length Extrapolation}},
  journal={arXiv preprint arXiv:2108.12409},
  year={2021}
}

@article{chen2023extending,
  author={Chen, Shouyuan and Wong, Sherman and Chen, Liangjian and Tian, Yuandong},
  title={{Extending Context Window of Large Language Models via Positional Interpolation}},
  journal={arXiv preprint arXiv:2306.15595},
  year={2023}
}

@article{peng2023yarn,
  author={Peng, Bowen and Quesnelle, Jeffrey and Fan, Honglu and Shippole, Enrico},
  title={{YaRN: Efficient Context Window Extension of Large Language Models}},
  journal={arXiv preprint arXiv:2309.00071},
  year={2023}
}

@article{yang2024qwen25,
  author={Yang, An and Yang, Baosong and Zhang, Beichen and Hui, Binyuan and others},
  title={{Qwen2.5 Technical Report}},
  journal={arXiv preprint arXiv:2412.15115},
  year={2024}
}

@article{grattafiori2024llama,
  author={Grattafiori, Aaron and Dubey, Abhimanyu and Jauhri, Abhinav and Pandey, Abhinav and Kadian, Abhishek and Al-Dahle, Ahmad and others},
  title={{The Llama 3 Herd of Models}},
  journal={arXiv preprint arXiv:2407.21783},
  year={2024}
}

@article{walsh2024olmo,
  author={Walsh, Pete and Soldaini, Luca and Groeneveld, Dirk and Lo, Kyle and Arora, Shane and others},
  title={{2 OLMo 2 Furious}},
  journal={arXiv preprint arXiv:2501.00656},
  year={2024}
}

@article{allal2025smollm2,
  author={Allal, Loubna Ben and Lozhkov, Anton and Bakouch, Elie and Blázquez, Gabriel Martín and Penedo, Guilherme and Tunstall, Lewis and others},
  title={{SmolLM2: When Smol Goes Big: Data-Centric Training of a Small Language Model}},
  journal={arXiv preprint arXiv:2502.02737},
  year={2025}
}

@article{lewis2020retrievalaugmented,
  author={Lewis, Patrick and Perez, Ethan and Piktus, Aleksandra and Petroni, Fabio and Karpukhin, Vladimir and Goyal, Naman and others},
  title={{Retrieval-Augmented Generation for Knowledge-Intensive NLP Tasks}},
  journal={arXiv preprint arXiv:2005.11401},
  year={2020}
}

@article{bai2023longbench,
  author={Bai, Yushi and Lv, Xin and Zhang, Jiajie and Lyu, Hongchang and Tang, Jiankai and Huang, Zhidian and others},
  title={{LongBench: A Bilingual, Multitask Benchmark for Long Context Understanding}},
  journal={arXiv preprint arXiv:2308.14508},
  year={2023}
}

@article{zhang2024inftybench,
  author={Zhang, Xinrong and Chen, Yingfa and Hu, Shengding and Xu, Zihang and Chen, Junhao and Hao, Moo Khai and others},
  title={{$\infty$Bench: Extending Long Context Evaluation Beyond 100K Tokens}},
  journal={arXiv preprint arXiv:2402.13718},
  year={2024}
}

@article{an2023leval,
  author={An, Chenxin and Gong, Shansan and Zhong, Ming and Zhao, Xingjian and Li, Mukai and Zhang, Jun and others},
  title={{L-Eval: Instituting Standardized Evaluation for Long Context Language Models}},
  journal={arXiv preprint arXiv:2307.11088},
  year={2023}
}

@article{shaham2023zeroscrolls,
  author={Shaham, Uri and Ivgi, Maor and Efrat, Avia and Berant, Jonathan and Levy, Omer},
  title={{ZeroSCROLLS: A Zero-Shot Benchmark for Long Text Understanding}},
  journal={arXiv preprint arXiv:2305.14196},
  year={2023}
}

@article{vodrahalli2024michelangelo,
  author={Vodrahalli, Kiran and Ontanon, Santiago and Tripuraneni, Nilesh and Xu, Kelvin and Jain, Sanil and Shivanna, Rakesh and others},
  title={{Michelangelo: Long Context Evaluations Beyond Haystacks via Latent Structure Queries}},
  journal={arXiv preprint arXiv:2409.12640},
  year={2024}
}

@article{kuratov2024babilong,
  author={Kuratov, Yuri and Bulatov, Aydar and Anokhin, Petr and Rodkin, Ivan and Sorokin, Dmitry and Sorokin, Artyom and others},
  title={{BABILong: Testing the Limits of LLMs with Long Context Reasoning-in-a-Haystack}},
  journal={arXiv preprint arXiv:2406.10149},
  year={2024}
}

@article{yen2024helmet,
  author={Yen, Howard and Gao, Tianyu and Hou, Minmin and Ding, Ke and Fleischer, Daniel and Izsak, Peter and others},
  title={{HELMET: How to Evaluate Long-Context Language Models Effectively and Thoroughly}},
  journal={arXiv preprint arXiv:2410.02694},
  year={2024}
}

@article{bai2024longbench,
  author={Bai, Yushi and Tu, Shangqing and Zhang, Jiajie and Peng, Hao and Wang, Xiaozhi and Lv, Xin and others},
  title={{LongBench v2: Towards Deeper Understanding and Reasoning on Realistic Long-context Multitasks}},
  journal={arXiv preprint arXiv:2412.15204},
  year={2024}
}

@article{olsson2022incontext,
  author={Olsson, Catherine and Elhage, Nelson and Nanda, Neel and Joseph, Nicholas and DasSarma, Nova and Henighan, Tom and others},
  title={{In-context Learning and Induction Heads}},
  journal={arXiv preprint arXiv:2209.11895},
  year={2022}
}

@article{hahn2019theoretical,
  author={Hahn, Michael},
  title={{Theoretical Limitations of Self-Attention in Neural Sequence Models}},
  journal={arXiv preprint arXiv:1906.06755},
  year={2019}
}

@article{longpre2021entitybased,
  author={Longpre, Shayne and Perisetla, Kartik and Chen, Anthony and Ramesh, Nikhil and DuBois, Chris and Singh, Sameer},
  title={{Entity-Based Knowledge Conflicts in Question Answering}},
  journal={arXiv preprint arXiv:2109.05052},
  year={2021}
}

@article{chen2022rich,
  author={Chen, Hung-Ting and Zhang, Michael J. Q. and Choi, Eunsol},
  title={{Rich Knowledge Sources Bring Complex Knowledge Conflicts: Recalibrating Models to Reflect Conflicting Evidence}},
  journal={arXiv preprint arXiv:2210.13701},
  year={2022}
}

@article{xie2023adaptive,
  author={Xie, Jian and Zhang, Kai and Chen, Jiangjie and Lou, Renze and Su, Yu},
  title={{Adaptive Chameleon or Stubborn Sloth: Revealing the Behavior of Large Language Models in Knowledge Conflicts}},
  journal={arXiv preprint arXiv:2305.13300},
  year={2023}
}

@article{wu2024clasheval,
  author={Wu, Kevin and Wu, Eric and Zou, James},
  title={{ClashEval: Quantifying the tug-of-war between an LLM's internal prior and external evidence}},
  journal={arXiv preprint arXiv:2404.10198},
  year={2024}
}

@article{cuconasu2024power,
  author={Cuconasu, Florin and Trappolini, Giovanni and Siciliano, Federico and Filice, Simone and Campagnano, Cesare and Maarek, Yoelle and others},
  title={{The Power of Noise: Redefining Retrieval for RAG Systems}},
  journal={arXiv preprint arXiv:2401.14887},
  year={2024}
}

@article{yoran2023making,
  author={Yoran, Ori and Wolfson, Tomer and Ram, Ori and Berant, Jonathan},
  title={{Making Retrieval-Augmented Language Models Robust to Irrelevant Context}},
  journal={arXiv preprint arXiv:2310.01558},
  year={2023}
}

@article{jia2017adversarial,
  author={Jia, Robin and Liang, Percy},
  title={{Adversarial Examples for Evaluating Reading Comprehension Systems}},
  journal={arXiv preprint arXiv:1707.07328},
  year={2017}
}

@article{xiao2023efficient,
  author={Xiao, Guangxuan and Tian, Yuandong and Chen, Beidi and Han, Song and Lewis, Mike},
  title={{Efficient Streaming Language Models with Attention Sinks}},
  journal={arXiv preprint arXiv:2309.17453},
  year={2023}
}

@article{gu2024attention,
  author={Gu, Xiangming and Pang, Tianyu and Du, Chao and Liu, Qian and Zhang, Fengzhuo and Du, Cunxiao and others},
  title={{When Attention Sink Emerges in Language Models: An Empirical View}},
  journal={arXiv preprint arXiv:2410.10781},
  year={2024}
}

@article{sun2024massive,
  author={Sun, Mingjie and Chen, Xinlei and Kolter, J. Zico and Liu, Zhuang},
  title={{Massive Activations in Large Language Models}},
  journal={arXiv preprint arXiv:2402.17762},
  year={2024}
}

@article{barbero2025llms,
  author={Barbero, Federico and Arroyo, Álvaro and Gu, Xiangming and Perivolaropoulos, Christos and Bronstein, Michael and Veličković, Petar and others},
  title={{Why do LLMs attend to the first token?}},
  journal={arXiv preprint arXiv:2504.02732},
  year={2025}
}

@article{wu2024retrieval,
  author={Wu, Wenhao and Wang, Yizhong and Xiao, Guangxuan and Peng, Hao and Fu, Yao},
  title={{Retrieval Head Mechanistically Explains Long-Context Factuality}},
  journal={arXiv preprint arXiv:2404.15574},
  year={2024}
}

@article{zhang2025queryfocused,
  author={Zhang, Wuwei and Yin, Fangcong and Yen, Howard and Chen, Danqi and Ye, Xi},
  title={{Query-Focused Retrieval Heads Improve Long-Context Reasoning and Re-ranking}},
  journal={arXiv preprint arXiv:2506.09944},
  year={2025}
}

@article{ye2024differential,
  author={Ye, Tianzhu and Dong, Li and Xia, Yuqing and Sun, Yutao and Zhu, Yi and Huang, Gao and others},
  title={{Differential Transformer}},
  journal={arXiv preprint arXiv:2410.05258},
  year={2024}
}

@article{qiu2025gated,
  author={Qiu, Zihan and Wang, Zekun and Zheng, Bo and Huang, Zeyu and Wen, Kaiyue and Yang, Songlin and others},
  title={{Gated Attention for Large Language Models: Non-linearity, Sparsity, and Attention-Sink-Free}},
  journal={arXiv preprint arXiv:2505.06708},
  year={2025}
}

@article{leviathan2024selective,
  author={Leviathan, Yaniv and Kalman, Matan and Matias, Yossi},
  title={{Selective Attention Improves Transformer}},
  journal={arXiv preprint arXiv:2410.02703},
  year={2024}
}

@article{lin2025forgetting,
  author={Lin, Zhixuan and Nikishin, Evgenii and He, Xu Owen and Courville, Aaron},
  title={{Forgetting Transformer: Softmax Attention with a Forget Gate}},
  journal={arXiv preprint arXiv:2503.02130},
  year={2025}
}

@article{tan2024scaling,
  author={Tan, Shawn and Yang, Songlin and Courville, Aaron and Panda, Rameswar and Shen, Yikang},
  title={{Scaling Stick-Breaking Attention: An Efficient Implementation and In-depth Study}},
  journal={arXiv preprint arXiv:2410.17980},
  year={2024}
}

@article{mohtashami2023landmark,
  author={Mohtashami, Amirkeivan and Jaggi, Martin},
  title={{Landmark Attention: Random-Access Infinite Context Length for Transformers}},
  journal={arXiv preprint arXiv:2305.16300},
  year={2023}
}

@article{wu2022memorizing,
  author={Wu, Yuhuai and Rabe, Markus N. and Hutchins, DeLesley and Szegedy, Christian},
  title={{Memorizing Transformers}},
  journal={arXiv preprint arXiv:2203.08913},
  year={2022}
}

@article{agarwal2024manyshot,
  author={Agarwal, Rishabh and Singh, Avi and Zhang, Lei M. and Bohnet, Bernd and Rosias, Luis and Chan, Stephanie and others},
  title={{Many-Shot In-Context Learning}},
  journal={arXiv preprint arXiv:2404.11018},
  year={2024}
}

@article{ding2024longrope,
  author={Ding, Yiran and Zhang, Li Lyna and Zhang, Chengruidong and Xu, Yuanyuan and Shang, Ning and Xu, Jiahang and others},
  title={{LongRoPE: Extending LLM Context Window Beyond 2 Million Tokens}},
  journal={arXiv preprint arXiv:2402.13753},
  year={2024}
}

@article{hsieh2024found,
  author={Hsieh, Cheng-Yu and Chuang, Yung-Sung and Li, Chun-Liang and Wang, Zifeng and Le, Long T. and Kumar, Abhishek and others},
  title={{Found in the Middle: Calibrating Positional Attention Bias Improves Long Context Utilization}},
  journal={arXiv preprint arXiv:2406.16008},
  year={2024}
}

@article{zhang2024found,
  author={Zhang, Zhenyu and Chen, Runjin and Liu, Shiwei and Yao, Zhewei and Ruwase, Olatunji and Chen, Beidi and others},
  title={{Found in the Middle: How Language Models Use Long Contexts Better via Plug-and-Play Positional Encoding}},
  journal={arXiv preprint arXiv:2403.04797},
  year={2024}
}

@article{an2024does,
  author={An, Chenxin and Zhang, Jun and Zhong, Ming and Li, Lei and Gong, Shansan and Luo, Yao and others},
  title={{Why Does the Effective Context Length of LLMs Fall Short?}},
  journal={arXiv preprint arXiv:2410.18745},
  year={2024}
}

@article{xiao2025mitigating,
  author={Xiao, Zikai and Wang, Ziyang and Ma, Wen and Zhang, Yan and Shen, Wei and Wang, Yan and others},
  title={{Mitigating Posterior Salience Attenuation in Long-Context LLMs with Positional Contrastive Decoding}},
  journal={arXiv preprint arXiv:2506.08371},
  year={2025}
}

@article{wang2024eliminating,
  author={Wang, Ziqi and Zhang, Hanlin and Li, Xiner and Huang, Kuan-Hao and Han, Chi and Ji, Shuiwang and others},
  title={{Eliminating Position Bias of Language Models: A Mechanistic Approach}},
  journal={arXiv preprint arXiv:2407.01100},
  year={2024}
}

@article{peysakhovich2023attention,
  author={Peysakhovich, Alexander and Lerer, Adam},
  title={{Attention Sorting Combats Recency Bias In Long Context Language Models}},
  journal={arXiv preprint arXiv:2310.01427},
  year={2023}
}

@article{yang2025rope,
  author={Yang, Bowen and Venkitesh, Bharat and Talupuru, Dwarak and Lin, Hangyu and Cairuz, David and Blunsom, Phil and others},
  title={{Rope to Nope and Back Again: A New Hybrid Attention Strategy}},
  journal={arXiv preprint arXiv:2501.18795},
  year={2025}
}

@article{gelberg2025extending,
  author={Gelberg, Yoav and Eguchi, Koshi and Akiba, Takuya and Cetin, Edoardo},
  title={{Extending the Context of Pretrained LLMs by Dropping Their Positional Embeddings}},
  journal={arXiv preprint arXiv:2512.12167},
  year={2025}
}

@article{kazemnejad2023impact,
  author={Kazemnejad, Amirhossein and Padhi, Inkit and Ramamurthy, Karthikeyan Natesan and Das, Payel and Reddy, Siva},
  title={{The Impact of Positional Encoding on Length Generalization in Transformers}},
  journal={arXiv preprint arXiv:2305.19466},
  year={2023}
}

@article{wu2025emergence,
  author={Wu, Xinyi and Wang, Yifei and Jegelka, Stefanie and Jadbabaie, Ali},
  title={{On the Emergence of Position Bias in Transformers}},
  journal={arXiv preprint arXiv:2502.01951},
  year={2025}
}

@article{chowdhury2026lost,
  author={Chowdhury, Borun D},
  title={{Lost in the Middle at Birth: An Exact Theory of Transformer Position Bias}},
  journal={arXiv preprint arXiv:2603.10123},
  year={2026}
}

@article{lu2024insights,
  author={Lu, Taiming and Gao, Muhan and Yu, Kuai and Byerly, Adam and Khashabi, Daniel},
  title={{Insights into LLM Long-Context Failures: When Transformers Know but Don't Tell}},
  journal={arXiv preprint arXiv:2406.14673},
  year={2024}
}

@article{bianchi2025hidden,
  author={Bianchi, Owen and Koretsky, Mathew J. and Willey, Maya and Alvarado, Chelsea X. and Nayak, Tanay and Asija, Adi and others},
  title={{Hidden in the Haystack: Smaller Needles are More Difficult for LLMs to Find}},
  journal={arXiv preprint arXiv:2505.18148},
  year={2025}
}

@article{dai2024deniahl,
  author={Dai, Hui and Pechi, Dan and Yang, Xinyi and Banga, Garvit and Mantri, Raghav},
  title={{DENIAHL: In-Context Features Influence LLM Needle-In-A-Haystack Abilities}},
  journal={arXiv preprint arXiv:2411.19360},
  year={2024}
}

@article{zou2024manyshot,
  author={Zou, Kaijian and Khalifa, Muhammad and Wang, Lu},
  title={{On Many-Shot In-Context Learning for Long-Context Evaluation}},
  journal={arXiv preprint arXiv:2411.07130},
  year={2024}
}

@article{wang2025adaptive,
  author={Wang, Yanbo and Xu, Zixiang and Huang, Yue and Gao, Chujie and Wu, Siyuan and Ye, Jiayi and others},
  title={{Adaptive Distraction: Probing LLM Contextual Robustness with Automated Tree Search}},
  journal={arXiv preprint arXiv:2502.01609},
  year={2025}
}

@article{yang2025llm,
  author={Yang, Minglai and Huang, Ethan and Zhang, Liang and Surdeanu, Mihai and Wang, William and Pan, Liangming},
  title={{How Is LLM Reasoning Distracted by Irrelevant Context? An Analysis Using a Controlled Benchmark}},
  journal={arXiv preprint arXiv:2505.18761},
  year={2025}
}

@article{wu2024reducing,
  author={Wu, Zijun and Liu, Bingyuan and Yan, Ran and Chen, Lei and Delteil, Thomas},
  title={{Reducing Distraction in Long-Context Language Models by Focused Learning}},
  journal={arXiv preprint arXiv:2411.05928},
  year={2024}
}

@article{shim2025beyond,
  author={Shim, Seongwoong and Kim, Myunsoo and Cho, Jae Hyeon and Lee, Byung-Jun},
  title={{Beyond RAG vs. Long-Context: Learning Distraction-Aware Retrieval for Efficient Knowledge Grounding}},
  journal={arXiv preprint arXiv:2509.21865},
  year={2025}
}

@article{li2025haystack,
  author={Li, Mufei and Fu, Dongqi and Wang, Limei and Zhang, Si and Zeng, Hanqing and Sancak, Kaan and others},
  title={{Haystack Engineering: Context Engineering for Heterogeneous and Agentic Long-Context Evaluation}},
  journal={arXiv preprint arXiv:2510.07414},
  year={2025}
}

@article{xia2026diagnosing,
  author={Xia, Shijie and Wang, Yikun and Huang, Zhen and Liu, Pengfei},
  title={{Diagnosing and Mitigating Context Rot in Long-horizon Search}},
  journal={arXiv preprint arXiv:2606.29718},
  year={2026}
}

@article{arjmandi2026distractoraware,
  author={Arjmandi, Mohsen},
  title={{Distractor-Aware Truncation: Disentangling Context-Length Effects from Signal Loss in Long-Context LLM Benchmarks}},
  journal={arXiv preprint arXiv:2608.03297},
  year={2026}
}

@article{wang2026intelligence,
  author={Wang, Weiwei and Min, Jiyong and Zou, Weijie},
  title={{Intelligence Degradation in Long-Context LLMs: Critical Threshold Determination via Natural Length Distribution Analysis}},
  journal={arXiv preprint arXiv:2601.15300},
  year={2026}
}

@article{du2026rope,
  author={Du, Yufeng and Harris, Phillip and Tian, Minyang and Huerta, Eliu A and Ronanki, Srikanth and Rongali, Subendhu and others},
  title={{RoPE Distinguishes Neither Positions Nor Tokens in Long Contexts, Provably}},
  journal={arXiv preprint arXiv:2605.15514},
  year={2026}
}

@article{vasylenko2025longcontext,
  author={Vasylenko, Pavlo and Pitorro, Hugo and Martins, André F. T. and Treviso, Marcos},
  title={{Long-Context Generalization with Sparse Attention}},
  journal={arXiv preprint arXiv:2506.16640},
  year={2025}
}

@article{bansal2025not,
  author={Bansal, Rachit and Zhang, Aston and Tiwari, Rishabh and Madaan, Lovish and Duvvuri, Sai Surya and Khatri, Devvrit and others},
  title={{Let's (not) just put things in Context: Test-Time Training for Long-Context LLMs}},
  journal={arXiv preprint arXiv:2512.13898},
  year={2025}
}

@article{ye2026dysco,
  author={Ye, Xi and Zhang, Wuwei and Yin, Fangcong and Yen, Howard and Chen, Danqi},
  title={{DySCO: Dynamic Attention-Scaling Decoding for Long-Context Language Models}},
  journal={arXiv preprint arXiv:2602.22175},
  year={2026}
}

@article{pei2026focusft,
  author={Pei, Zehua and Zhen, Hui-Ling and Yu, Xianzhi and Pan, Sinno Jialin and Yuan, Mingxuan and Yu, Bei},
  title={{FocuSFT: Bilevel Optimization for Dilution-Aware Long-Context Fine-Tuning}},
  journal={arXiv preprint arXiv:2605.09932},
  year={2026}
}

@article{bui2026make,
  author={Bui, Ngoc and Nguyen, Hieu Trung and Cohan, Arman and Ying, Rex},
  title={{Make Each Token Count: Towards Improving Long-Context Performance with KV Cache Eviction}},
  journal={arXiv preprint arXiv:2605.09649},
  year={2026}
}

@article{huang2026threshold,
  author={Huang, Xingyue and Ding, Xueying and Ju, Mingxuan and Liu, Yozen and Shah, Neil and Zhao, Tong},
  title={{Threshold Differential Attention for Sink-Free, Ultra-Sparse, and Non-Dispersive Language Modeling}},
  journal={arXiv preprint arXiv:2601.12145},
  year={2026},
  note={ACL 2026}
}

@article{gu2026long,
  author={Gu, Shangding},
  title={{Long Context, Less Focus: A Scaling Gap in LLMs Revealed through Privacy and Personalization}},
  journal={arXiv preprint arXiv:2602.15028},
  year={2026}
}

@article{mudarisov2025limitations,
  author={Mudarisov, Timur and Burtsev, Mikhail and Petrova, Tatiana and State, Radu},
  title={{Limitations of Normalization in Attention Mechanism}},
  journal={arXiv preprint arXiv:2508.17821},
  year={2025}
}

@article{chen2025critical,
  author={Chen, Shi and Lin, Zhengjiang and Polyanskiy, Yury and Rigollet, Philippe},
  title={{Critical attention scaling in long-context transformers}},
  journal={arXiv preprint arXiv:2510.05554},
  year={2025},
  note={ICLR 2026}
}

@article{anson2025scaleinvariant,
  author={Anson, Ben and Wang, Xi and Aitchison, Laurence},
  title={{Scale-invariant Attention}},
  journal={arXiv preprint arXiv:2505.17083},
  year={2025}
}

@article{li2025information,
  author={Li, Kewei and Kong, Yanwen and Xu, Yiping and Su, Jianlin and Huang, Lan and Zhang, Ruochi and others},
  title={{Information Entropy Invariance: Enhancing Length Extrapolation in Attention Mechanisms}},
  journal={arXiv preprint arXiv:2501.08570},
  year={2025}
}

@article{zhang2024selective,
  author={Zhang, Xuechen and Chang, Xiangyu and Li, Mingchen and Roy-Chowdhury, Amit and Chen, Jiasi and Oymak, Samet},
  title={{Selective Attention: Enhancing Transformer through Principled Context Control}},
  journal={arXiv preprint arXiv:2411.12892},
  year={2024}
}

@article{lee2025seal,
  author={Lee, Changhun and Seok, Minsang and Jin, Jun-gyu and Cho, Younghyun and Park, Eunhyeok},
  title={{SEAL: Scaling to Emphasize Attention for Long-Context Retrieval}},
  journal={arXiv preprint arXiv:2501.15225},
  year={2025}
}

@article{chirkova2025provence,
  author={Chirkova, Nadezhda and Formal, Thibault and Nikoulina, Vassilina and Clinchant, Stéphane},
  title={{Provence: efficient and robust context pruning for retrieval-augmented generation}},
  journal={arXiv preprint arXiv:2501.16214},
  year={2025}
}

@article{choi2025think,
  author={Choi, Daewon and Lee, Jimin and Tack, Jihoon and Song, Woomin and Dingliwal, Saket and Jayanthi, Sai Muralidhar and others},
  title={{Think Clearly: Improving Reasoning via Redundant Token Pruning}},
  journal={arXiv preprint arXiv:2507.08806},
  year={2025}
}

@article{xu2024knowledge,
  author={Xu, Rongwu and Qi, Zehan and Guo, Zhijiang and Wang, Cunxiang and Wang, Hongru and Zhang, Yue and others},
  title={{Knowledge Conflicts for LLMs: A Survey}},
  journal={arXiv preprint arXiv:2403.08319},
  year={2024}
}

@article{wang2025unable,
  author={Wang, Chupei and Sun, Jiaqiu Vince},
  title={{Unable to Forget: Proactive Interference Reveals Working Memory Limits in LLMs Beyond Context Length}},
  journal={arXiv preprint arXiv:2506.08184},
  year={2025}
}

@article{du2025context,
  author={Du, Yufeng and Tian, Minyang and Ronanki, Srikanth and Rongali, Subendhu and Bodapati, Sravan and Galstyan, Aram and others},
  title={{Context Length Alone Hurts LLM Performance Despite Perfect Retrieval}},
  journal={arXiv preprint arXiv:2510.05381},
  year={2025}
}

@article{gao2026first,
  author={Gao, Muhan and Chen, Zih-Ching and Huang, Kuan-Hao},
  title={{The First Drop of Ink: Nonlinear Impact of Distracting Information in Long-Context Reasoning}},
  journal={arXiv preprint arXiv:2605.10828},
  year={2026}
}

@article{ghaffari2026context,
  author={Ghaffari, Meysam and Fatehi, Nina and Sen, Bhaskar and Sabetpour, Nasim and Morato, Carlos},
  title={{Context Poisoning as Extreme-Value Attention Interference in Long-Context Language Models}},
  journal={arXiv preprint arXiv:2609.22101},
  year={2026}
}

@article{repantis2026separating,
  author={Repantis, Vyzantinos and Gawde, Ameya and Singh, Harshvardhan and Alekar, Rohit and Zhang, Cien and Karslioglu, Svetlana and others},
  title={{Separating Semantic Competition from Context Length in RAG Reading}},
  journal={arXiv preprint arXiv:2605.27294},
  year={2026}
}

@article{lin2026beyond,
  author={Lin, Tianwei and Zhou, Zuyi and Zhao, Xinda and Wang, Chenke and Li, Xiaohong and Chen, Yu and others},
  title={{Beyond the Needle's Illusion: Decoupled Evaluation of Evidence Access and Use under Semantic Interference at 326M-Token Scale}},
  journal={arXiv preprint arXiv:2601.20276},
  year={2026}
}

@article{levy2024same,
  author={Levy, Mosh and Jacoby, Alon and Goldberg, Yoav},
  title={{Same Task, More Tokens: the Impact of Input Length on the Reasoning Performance of Large Language Models}},
  journal={arXiv preprint arXiv:2402.14848},
  year={2024}
}

@article{shi2023large,
  author={Shi, Freda and Chen, Xinyun and Misra, Kanishka and Scales, Nathan and Dohan, David and Chi, Ed and others},
  title={{Large Language Models Can Be Easily Distracted by Irrelevant Context}},
  journal={arXiv preprint arXiv:2302.00093},
  year={2023}
}

@article{tworkowski2023focused,
  author={Tworkowski, Szymon and Staniszewski, Konrad and Pacek, Mikołaj and Wu, Yuhuai and Michalewski, Henryk and Miłoś, Piotr},
  title={{Focused Transformer: Contrastive Training for Context Scaling}},
  journal={arXiv preprint arXiv:2307.03170},
  year={2023}
}

@article{velickovic2024softmax,
  author={Veličković, Petar and Perivolaropoulos, Christos and Barbero, Federico and Pascanu, Razvan},
  title={{Softmax is not Enough (for Sharp Size Generalisation)}},
  journal={arXiv preprint arXiv:2410.01104},
  year={2024}
}

@article{nakanishi2025scalablesoftmax,
  author={Nakanishi, Ken M.},
  title={{Scalable-Softmax Is Superior for Attention}},
  journal={arXiv preprint arXiv:2501.19399},
  year={2025}
}

@article{chiang2022overcoming,
  author={Chiang, David and Cholak, Peter},
  title={{Overcoming a Theoretical Limitation of Self-Attention}},
  journal={arXiv preprint arXiv:2202.12172},
  year={2022}
}

@article{liu2023lost,
  author={Liu, Nelson F. and Lin, Kevin and Hewitt, John and Paranjape, Ashwin and Bevilacqua, Michele and Petroni, Fabio and others},
  title={{Lost in the Middle: How Language Models Use Long Contexts}},
  journal={arXiv preprint arXiv:2307.03172},
  year={2023}
}

@article{hsieh2024ruler,
  author={Hsieh, Cheng-Ping and Sun, Simeng and Kriman, Samuel and Acharya, Shantanu and Rekesh, Dima and Jia, Fei and others},
  title={{RULER: What's the Real Context Size of Your Long-Context Language Models?}},
  journal={arXiv preprint arXiv:2404.06654},
  year={2024}
}

@article{modarressi2025nolima,
  author={Modarressi, Ali and Deilamsalehy, Hanieh and Dernoncourt, Franck and Bui, Trung and Rossi, Ryan A. and Yoon, Seunghyun and others},
  title={{NoLiMa: Long-Context Evaluation Beyond Literal Matching}},
  journal={arXiv preprint arXiv:2502.05167},
  year={2025}
}

@article{wang2024length,
  author={Wang, Jie and Ji, Tao and Wu, Yuanbin and Yan, Hang and Gui, Tao and Zhang, Qi and others},
  title={{Length Generalization of Causal Transformers without Position Encoding}},
  journal={arXiv preprint arXiv:2404.12224},
  year={2024}
}

@article{su2026attention,
  author={Su, Zunhai and Zhang, Hengyuan and Wu, Wei and Zhang, Yifan and Liu, Yaxiu and Xiao, He and others},
  title={{Attention Sink in Transformers: A Survey on Utilization, Interpretation, and Mitigation}},
  journal={arXiv preprint arXiv:2604.10098},
  year={2026}
}

@article{ranmilo2026attention,
  author={Ran-Milo, Yuval},
  title={{Attention Sinks Are Provably Necessary in Softmax Transformers: Evidence from Trigger-Conditional Tasks}},
  journal={arXiv preprint arXiv:2603.11487},
  year={2026}
}

@article{rizwan2026new,
  author={Rizwan, Sara and Salam, Samaanah Abdus and Uddin, Mohammed Mudassir},
  title={{Do New Attention Mechanisms Actually Fix Attention Sinks at Million-Token Context?}},
  journal={arXiv preprint arXiv:2609.08574},
  year={2026}
}

@article{dongre2026attention,
  author={Dongre, Vardhan and Hsieh, Joseph and Lai, Viet Dac and Yoon, Seunghyun and Bui, Trung and Hakkani-Tür, Dilek},
  title={{When Attention Closes: How LLMs Lose the Thread in Multi-Turn Interaction}},
  journal={arXiv preprint arXiv:2605.12922},
  year={2026}
}

@misc{hong2025contextrot,
  author={Hong, Kelly and Troynikov, Anton and Huber, Jeff},
  title={{Context Rot: How Increasing Input Tokens Impacts {LLM} Performance}},
  howpublished={Chroma Technical Report, \url{https://research.trychroma.com/context-rot}},
  year={2025}
}

\end{document}